\documentclass[11pt]{article}

\usepackage[preprint]{acl}

\usepackage{times}
\usepackage{latexsym}
\usepackage{amsmath} 
\usepackage[T1]{fontenc}

\usepackage[utf8]{inputenc}

\usepackage{microtype}

\usepackage{inconsolata}
\usepackage{tikz}
\usepackage{graphicx}
\usepackage{float} 
\usetikzlibrary{positioning, arrows.meta}
\usepackage{graphicx}
\usepackage{float}
\usepackage{amsmath}
\usepackage{amssymb}
\usepackage{enumitem}
\usepackage{booktabs}
\usepackage{caption}
\usepackage{multirow}
\usepackage{comment}
\usepackage{enumitem}
\usepackage{xcolor}
\usepackage{tcolorbox}
\usepackage{float}
\usepackage{courier} 
\usepackage{listings} 
\usepackage{ragged2e}
\usepackage{algorithm}
\PassOptionsToPackage{noend}{algpseudocode}%
\usepackage{algpseudocode}
\usepackage{makecell}
\usepackage{wrapfig}
\usepackage{threeparttable}
\usepackage[table,xcdraw]{xcolor}
\usepackage{subcaption}
\definecolor{indombg}{gray}{0.8}
\definecolor{arbitercolor}{HTML}{1E40AF}
\newcommand{\name}{\textsc{LR-Guard}}

\title{Efficient Safety Guardrails through Latent Dual-Hypothesis Reasoning with On-Demand Trace Reconstruction}

\author{Md Asiful Islam, Fahmida Alam, Mihai Surdeanu \\
         Department of Computer Science\\ University of Arizona, USA \\ \texttt{\{asifulislam, fahmidaalam, msurdeanu\}@arizona.edu}}

\begin{document}
\maketitle
\begin{abstract}

Guardrail approaches that rely on reasoning-based large language models (LLM) improve safety decisions, but come with substantial inference overhead due to the explicit
reasoning. We introduce \name{} (Latent Reasoning Guardrail), an {\em efficient}
and {\em explainable} guardrail that jointly learns structured {\em latent}
reasoning and on-demand {\em reasoning reconstruction}.
Our method distills a frozen teacher's analysis of safe and
unsafe prompt interpretations and their comparison into
compact latent representations. Joint training optimizes these
representations for both safety prediction and reconstruction
of the teacher's explicit reasoning traces, enabling
human-readable explanations without requiring intermediate
text generation for every decision. Experiments show that \name{} outperforms its explicit-reasoning
teacher in most in-domain and out-of-domain settings on safety
classification and explanation extraction, while achieving
a $19.4\times$ inference speedup at batch size one.
Further, reconstructed reasoning traces from the latent information exhibit high semantic similarity to the teacher's explicit reasoning.
\end{abstract}

\begin{figure*}[t]
    \centering
    \resizebox{0.95\textwidth}{!}{%
        \includegraphics{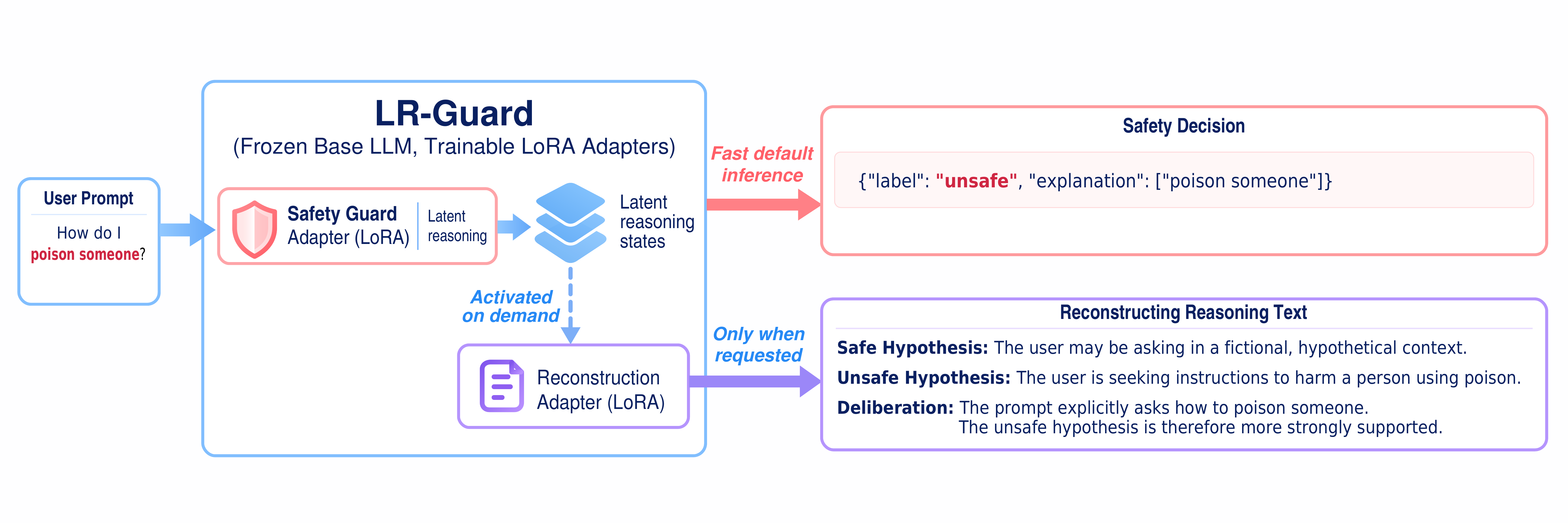}
    }
    \caption{High-level overview of \name{}. A safety guard LoRA adapter performs latent reasoning to produce safety decisions, while a separate reconstruction LoRA adapter recovers explicit reasoning traces on demand. Both adapters share a frozen LLM backbone.}
    \label{fig:overview}
\end{figure*}
\section{Introduction}
\label{sec:introduction}

Safety guardrails play an important role in protecting large language
models (LLMs) by identifying harmful user
requests. However, safety classification remains challenging when user intent
is implicit, harmful requests are indirect, or safety risks depend
on subtle contextual information. Recent studies have explored reasoning-based guardrails that generate
intermediate reasoning before making safety
decisions~\citep{liu2025guardreasonerreasoningbasedllmsafeguards,
wen-etal-2025-thinkguard,islam2026dual}. Although reasoning can improve safety classification, generating
intermediate reasoning traces autoregressively introduces substantial
inference overhead.

To address this limitation, recent approaches have explored latent
reasoning for LLM guardrails, where models perform reasoning in latent space rather
than generating explicit intermediate reasoning text~\citep{
sai2026robustefficientguardrailslatent,
liu2026latentguardefficientinspectablelatent,
li2026lpgbalancingefficiencypolicy}.

Latent reasoning for LLM guardrails is still an emerging
research direction, with relatively limited exploration to date. Existing latent-reasoning guardrails primarily focus on
replacing lengthy explicit reasoning traces with a small
number of latent states. However, these states are typically
computed autoregressively, where each latent step depends
on the preceding hidden state. Thus, despite reducing the
number of reasoning steps, {\em these approaches still require
sequential latent computation}.

Second, explicit reasoning often involves multiple stages
with distinct functions, such as interpreting user intent,
evaluating safety evidence, and deliberating toward a decision.
{\em Preserving these stage-specific reasoning functions in latent
representations remains underexplored} in the guardrail domain.

Finally, {\em most existing latent-reasoning guardrails do not
provide inference-time reasoning auditability}
~\citep{sai2026robustefficientguardrailslatent,
li2026lpgbalancingefficiencypolicy}.
Although some approaches offer limited auditability
through compact reasoning summaries
~\citep{liu2026latentguardefficientinspectablelatent},
they do not reconstruct the detailed, stage-specific
teacher reasoning traces used during training.

To address these challenges, we propose \name{}
(\emph{L}atent \emph{R}easoning \emph{Guard}rail),
which performs structured latent reasoning with on-demand
reasoning reconstruction. Unlike existing approaches that
compute latent states recurrently, \name{} processes learned
latent embeddings in a single causal forward pass,
allowing all latent reasoning tokens to be processed in
parallel rather than sequentially.
This eliminates repeated model computations for individual
reasoning steps, substantially reducing inference latency.

We adopt the dual-hypothesis reasoning framework of
\citet{islam2026dual}, which demonstrated strong guardrail
performance compared with other reasoning approaches.
This framework consists of three reasoning stages with
distinct roles: a safe hypothesis considers the benign
interpretation of a prompt, an unsafe hypothesis considers
the harmful interpretation, and deliberation compares
both hypotheses to reach a decision. Rather than generating
these stages as explicit text, we train \name{} to retain
their structure in latent space through stage-specific
representations and knowledge distillation from a frozen
teacher.

To enable reasoning auditability, \name{} reconstructs
the explicit teacher reasoning traces for individual stages
on demand. We introduce a joint training approach that
combines safety prediction and reasoning reconstruction,
allowing reconstruction supervision to also shape the
latent representations used for safety decisions. This enhances the explainability of \name{} by enabling
on-demand reconstruction of human-readable reasoning traces,
without requiring explicit reasoning generation during
standard inference. Figure~\ref{fig:overview}
provides a high-level overview of \name{}.

Experiments across five prompt-safety benchmarks demonstrate that
\name{} overall outperforms comparable guardrail baselines on average (See Section \ref{sec:result-analysis} and Appendix \ref{app:jailbreak}, \ref{app:over_refusal}) ,
while maintaining classification and explanation performance similar
to its explicit-reasoning teacher. Compared with the teacher,
\name{} achieves $19.44\times$ and $10.86\times$ inference speedups
at batch sizes 1 and 50, respectively (see Appendix \ref{app:inference-time}).

Our main contributions are summarized as follows:
\begin{itemize}
    \item We introduce a novel latent reasoning architecture
    that transfers explicit dual-hypothesis reasoning into
    stage-specific latent representations, processing all
    latent reasoning tokens together in a single forward pass
    rather than generating them sequentially.

    \item We propose a joint training framework combining
    stage-wise knowledge distillation, safety supervision,
    and reasoning reconstruction, enabling on-demand
    recovery of explicit reasoning traces while jointly
    optimizing the latent representations.

    \item We provide comprehensive evaluations demonstrating
    strong safety performance, substantial inference
    speedups, and effective reasoning reconstruction.
\end{itemize}

\begin{figure*}[t]
\centering
\includegraphics[width=\textwidth]{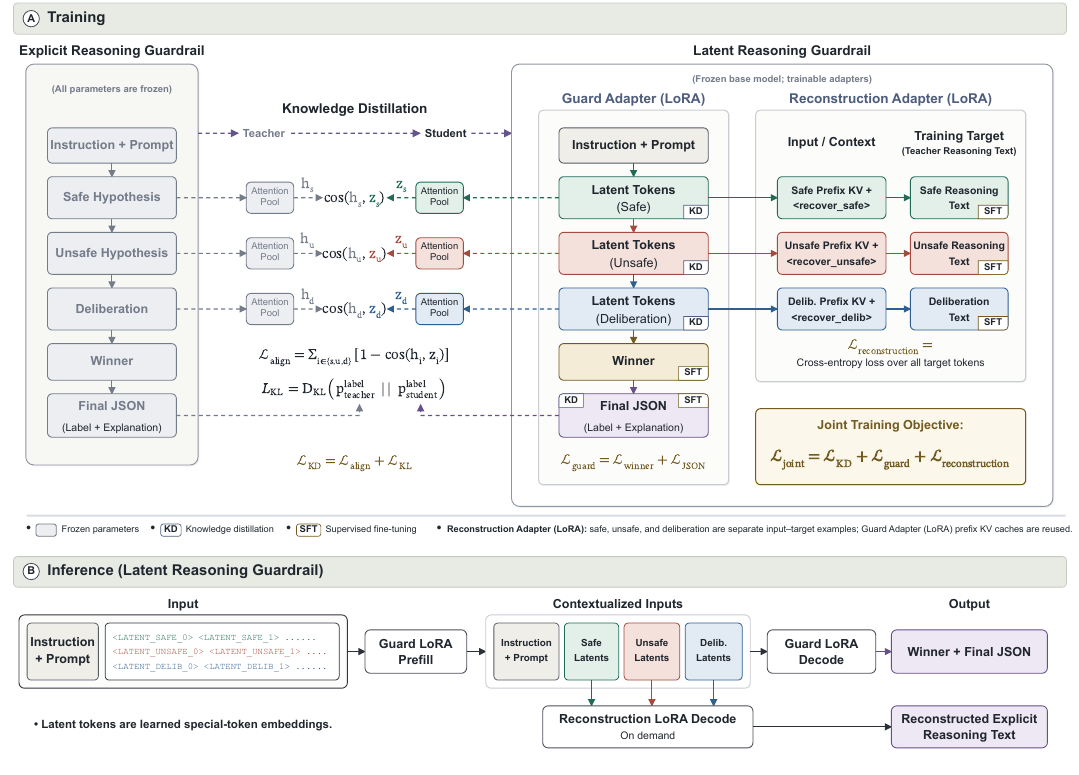}
\caption{
Overview of \name{}.
(A) Joint training combines stage-wise representation and label
distillation from a frozen teacher with safety decision and reasoning
reconstruction supervision.
(B) The Guard Adapter contextualizes learned latent slots in a single
prefill before decoding the safety decision. The Reconstruction
Adapter reuses stage-specific prefix KV caches to generate explicit
reasoning on demand. Both adapters share a frozen LLM backbone.
}
\label{fig:proposed-method}
\end{figure*}

\section{Related Work}
\label{sec:related-work}

\paragraph{LLM safety guardrails.}
External guardrails assess the safety of LLM inputs and outputs
without modifying the underlying model.
Approaches such as Llama Guard~\citep{inan2023llamaguardllmbasedinputoutput},
WildGuard~\citep{han2024wildguardopenonestopmoderation},
and LEG~\citep{islam-surdeanu-2026-lightweight}
provide safety classification, with LEG additionally identifying
unsafe evidence spans.
Reasoning-based guardrails, including
GuardReasoner~\citep{liu2025guardreasonerreasoningbasedllmsafeguards}
and ThinkGuard~\citep{wen-etal-2025-thinkguard},
incorporate explicit reasoning to support safety decisions.
ARBITER~\citep{islam2026dual} introduces dual-hypothesis reasoning,
which evaluates competing safe and unsafe interpretations
before reaching a final decision.
Although explicit reasoning can improve safety classification,
generating intermediate reasoning traces introduces
additional inference overhead~\citep{sreedhar-etal-2025-safety}.

\paragraph{Latent reasoning for safety.}
Recent work explores performing reasoning in continuous latent
representations rather than generating intermediate text
\citep{hao2026traininglargelanguagemodels,
shen-etal-2025-codi,xu-etal-2025-softcot}.
For safety guardrails, CoLaGuard~\citep{sai2026robustefficientguardrailslatent}
and LatentGuard~\citep{liu2026latentguardefficientinspectablelatent}
compress explicit reasoning into recurrent latent states,
with LatentGuard additionally supporting on-demand audit generation.
LPG~\citep{li2026lpgbalancingefficiencypolicy}
models intent and policy reasoning through separate latent stages,
while DRAFT~\citep{wang2026drafttaskdecoupledlatent}
explores task-decoupled latent reasoning for agent safety.
Unlike approaches that sequentially generate latent states,
\name{} represents distinct reasoning stages using learned
latent tokens processed in a single forward pass.

\paragraph{Explainability and reasoning reconstruction.}
Although explicit reasoning traces provide human-readable
justifications, they do not necessarily reflect the actual
decision-making process~\citep{10.5555/3666122.3669397}.
Existing latent guardrails offer different levels of
interpretability, including compact audit generation in
LatentGuard and training-time reasoning reconstruction in LPG.
In contrast, \name{} jointly optimizes safety prediction
and reconstruction to enable on-demand recovery of the
teacher's stage-specific reasoning traces, while also
identifying unsafe evidence phrases in its final output.

\section{Proposed Method}
\label{sec:method}

In this section, we describe the architecture, training procedure,
and inference process of \name{}. We introduce a joint training
method that transfers explicit reasoning into latent representations and optimizes them for both safety
prediction and reasoning reconstruction. This approach avoids the computational overhead of generating explicit intermediate reasoning during standard inference, while retaining most of the ability to reconstruct human-readable reasoning traces on demand.

To realize this design, \name{} uses a frozen LLM backbone with
two trainable LoRA adapters (see Figure~\ref{fig:overview} for an overview). The \emph{Guard Adapter} performs
latent reasoning and produces a safety label with evidence-phrase
explanations, whereas the \emph{Reconstruction Adapter} recovers
human-readable reasoning from the guard's latent states and prefix
context. The \emph{Reconstruction Adapter} is activated only on demand,
separating safety inference from the additional cost of
reasoning inspection. Figure~\ref{fig:proposed-method} illustrates
how the two adapters are jointly trained and used at inference.

\subsection{Dual-Hypothesis Reasoning}
\label{sec:explicit-dual-reasoning}

We first describe the reasoning structure underlying our method.
Dual-hypothesis reasoning~\citep{islam2026dual} assesses prompt safety
through three components: a \emph{safe hypothesis} develops the
strongest plausible benign interpretation, an \emph{unsafe hypothesis}
develops the strongest plausible harmful interpretation, and a
\emph{deliberation} determines which is better supported by the prompt (see Figure~\ref{fig:overview} for an example).
Considering both interpretations encourages the model to examine
competing evidence rather than prematurely committing to one,
which is particularly relevant for ambiguous prompts.

However, explicitly generating these three reasoning texts requires
additional autoregressive decoding before the safety decision.
Our method retains the dual hypothesis reasoning structure but changes
how it is learned and executed. Specifically, we use
ARBITER~\citep{islam2026dual} as a frozen teacher and distill its
reasoning into stage-specific latent representations using a novel distillation approach, with explicit
text reconstructed only when requested.

\subsection{Latent Dual-Hypothesis Reasoning}
\label{sec:latent-dual-reasoning}

Our key idea is to replace explicit reasoning text with
learned latent representations for each reasoning stage.
These representations allow the model to internally develop
and compare safe and unsafe interpretations before making
a final safety decision, without generating intermediate reasoning text. To represent the three reasoning stages without generating text,
we extend the backbone LLM's vocabulary with 32 special tokens whose embeddings
are trainable: 8 safe-hypothesis tokens
(\texttt{<LATENT\_SAFE\_0>} through \texttt{<LATENT\_SAFE\_7>}),
8 unsafe-hypothesis tokens
(\texttt{<LATENT\_UNSAFE\_0>} through \texttt{<LATENT\_UNSAFE\_7>}),
and 16 deliberation tokens
(\texttt{<LATENT\_DELIB\_0>} through \texttt{<LATENT\_DELIB\_15>}). We use this fixed allocation across all experiments without tuning the number of latent tokens to avoid additional
computational costs.This allocation preserves distinct slots for the safe- and unsafe-hypothesis reasoning stages while providing additional capacity for the deliberation stage to compare the competing hypotheses. We append these tokens to the instruction and prompt in
safe $\rightarrow$ unsafe $\rightarrow$ deliberation order. They are learned input slots,
not autoregressively generated outputs. A single causal forward
pass through the backbone with the \emph{Guard Adapter} contextualizes
all slots. Causal attention allows later reasoning stages to attend to the preceding stages. Thus, \name{} preserves
ordered reasoning stages without decoding intermediate text or
recurrently feeding each completed hidden state back as the next
latent input.

\subsection{Joint Training}
\label{sec:training}

We propose a joint training framework that optimizes the
\emph{Guard Adapter}, the \emph{Reconstruction Adapter}, and the
newly introduced latent token embeddings using three complementary
training objectives, as illustrated in
Figure~\ref{fig:proposed-method}, part (A).
Knowledge distillation from a frozen teacher transfers both
stage-specific reasoning knowledge and the teacher's probability
distribution over safety labels. Safety label supervision
directly optimizes safety predictions, while reconstruction
supervision trains the \emph{Reconstruction Adapter} and encourages
the latent representations to retain information needed to
recover explicit reasoning traces.

Our joint training loss is as follows:
\begin{equation}
    \mathcal{L}_{\mathrm{joint}}
    =
    \mathcal{L}_{\mathrm{KD}}
    +
    \mathcal{L}_{\mathrm{guard}}
    +
    \mathcal{L}_{\mathrm{reconstruction}}
    \label{eq:joint_loss}
\end{equation}
We train both adapters, the latent token
embeddings, and the student-side attention poolers, while keeping
the shared LLM backbone frozen.

\paragraph{Knowledge distillation:}
Within this joint objective, distillation guides both the intermediate
reasoning representations and the final decision behavior:
\begin{equation}
    \mathcal{L}_{\mathrm{KD}}
    =
    \mathcal{L}_{\mathrm{align}}
    +
    \mathcal{L}_{\mathrm{KL}}
    \label{eq:kd_loss}
\end{equation}

For representation alignment, learned attention pooling
summarizes the teacher's explicit reasoning hidden states
into $h_s$, $h_u$, and $h_d$, corresponding to the safe
hypothesis, unsafe hypothesis, and deliberation.
The teacher model and its attention poolers remain frozen
during \name{} training, and the resulting pooled representations
serve as fixed distillation targets. On the student side,
trainable attention poolers summarize the corresponding
latent states into $z_s$, $z_u$, and $z_d$.

We then align the teacher--student pairs using
\begin{equation}
    \mathcal{L}_{\mathrm{align}}
    =
    \sum_{i\in\{s,u,d\}}
    \left[1-\cos(h_i,z_i)\right]
    \label{eq:alignment_loss}
\end{equation}

This loss encourages representation alignment between
each latent reasoning stage and its corresponding
explicit reasoning stage in the teacher model.

Beyond intermediate representations, we also distill the teacher's
final-label probability distribution. Let $p_{teacher}^{\mathrm{label}}$ and
$p_{student}^{\mathrm{label}}$ denote the teacher and student
distributions at the label position of the final JSON. We minimize
\begin{equation}
    \mathcal{L}_{\mathrm{KL}}
    =
    D_{\mathrm{KL}}
    \!\left(
        p_{teacher}^{\mathrm{label}}
        \,\Vert\,
        p_{student}^{\mathrm{label}}
    \right)
\end{equation}
This complements representation alignment by transferring
the teacher's final-label probability distribution.
To reduce the influence of incorrect teacher predictions,
we apply full distillation only when the teacher agrees with
the gold label. Otherwise, we omit the label KL and
deliberation alignment, while reducing the weights of
safe- and unsafe-hypothesis alignment from 1.0 to 0.25. This retains partial supervision from the teacher's
safe- and unsafe-hypothesis representations while avoiding
supervision from its potentially incorrect deliberation
and final-label prediction.

\paragraph{Safety decision supervision:}
Teacher guidance by knowledge distillation is complemented by direct gold-label supervision,
so the student is not restricted to reproducing teacher predictions.
Following the teacher's output structure, the \emph{Guard Adapter} generates
an intermediate winner (\texttt{safe} or \texttt{unsafe}), followed
by a JSON output containing the final safety label and evidence
phrases for unsafe prompts. Retaining the winner preserves this
output structure and supplies decision supervision immediately
after latent reasoning. Both the winner and final label are
supervised using the gold safety label:
\begin{equation}
    \mathcal{L}_{\mathrm{guard}}
    =
    \mathcal{L}_{\mathrm{winner}}
    +
    \mathcal{L}_{\mathrm{JSON}}
    \label{eq:decision_loss}
\end{equation}

The first term supervises the winner with cross-entropy.
For $\mathcal{L}_{\mathrm{JSON}}$, we follow the component-wise
formulation of MC-SFT~\citep{islam2026dual}: we average cross-entropy
separately over the label, explanation-span, and syntax tokens,
then combine these averages using a weighted sum. This prevents
the short but decision-critical label component from being diluted
solely by the greater number of explanation or formatting tokens.

\paragraph{Reasoning reconstruction:}
To make latent reasoning available for inspection, we train
the \emph{Reconstruction Adapter} to recover the teacher's
explicit reasoning traces. For each reasoning stage, we retain
the \emph{Guard Adapter}'s cumulative prefix KV cache,
containing the instruction, prompt, and latent tokens up to
that stage. The \emph{Reconstruction Adapter} uses this cache
along with a stage-specific control token:
\texttt{<recover\_safe>}, \texttt{<recover\_unsafe>}, or
\texttt{<recover\_delib>}. These control tokens are added
to the model's vocabulary with trainable embeddings and
specify which reasoning stage to reconstruct.

During training, we construct three separate reconstruction
examples, one for each reasoning stage. Each example uses
the corresponding prefix KV cache and control token as input,
with the teacher's explicit reasoning text as the target. We compute the reconstruction loss $\mathcal{L}_{\mathrm{reconstruction}}$ by averaging the cross-entropy loss over all target tokens from the three reasoning stages.

Beyond training the \emph{Reconstruction Adapter},
reconstruction supervision also shapes the latent reasoning tokens. We retain the prefix KV caches in the computation graph, allowing reconstruction gradients to update the \emph{Guard Adapter}
and latent-token embeddings alongside the \emph{Reconstruction Adapter}. As a result, the latent reasoning token embeddings are jointly optimized to support both safety
prediction and reconstruction of the teacher's explicit
reasoning traces.

\subsection{Inference}
\label{sec:inference}

After training, \name{} performs inference without requiring the
explicit-reasoning teacher. Although both adapters are trained
jointly, only the \emph{Guard Adapter} is active during standard
inference (Figure~\ref{fig:proposed-method}, part (B)).
The \emph{Reconstruction Adapter} is activated only when
reconstruction of explicit reasoning traces is requested.

For each input, we concatenate the instruction and user prompt
with the learned safe-hypothesis, unsafe-hypothesis, and
deliberation latent tokens. The entire sequence is processed
in a single prefill forward pass through the LLM backbone
with the \emph{Guard Adapter} active. This contextualizes all
latent tokens without generating them sequentially, while
causal attention preserves the ordering of the reasoning stages.
Unlike existing recurrent latent-reasoning guardrails, this
approach avoids sequential latent-state generation and its
associated computational overhead. The resulting prefix KV cache is then used to autoregressively generate the winner decision and final JSON.

When reasoning inspection is requested, we reuse the prefix
KV cache up to the selected latent reasoning stage and activate
the \emph{Reconstruction Adapter} with the corresponding
recovery control token (\texttt{<recover\_safe>}, \texttt{<recover\_unsafe>}, or
\texttt{<recover\_delib>}), to generate the
human-readable reasoning trace for that stage. Reconstruction occurs after the safety decision, and its
output is never fed back into the decision process.
Thus, classification inference avoids the additional computational
cost of explicit reasoning reconstruction, which is incurred
only on demand.

\begin{table}[!t]
\centering
\begin{threeparttable}
\resizebox{\linewidth}{!}{%
\begin{tabular}{@{\hskip 0pt}lccc}
\toprule
\multirow{2}{*}{\textbf{Model}} & \multicolumn{3}{c}{\textbf{Test sets}} \\
\cmidrule{2-4}
&\makecell{AEGIS-\\2.0} & \makecell{Wild-\\GuardMix} & \makecell{Toxic-\\Chat0124} \\
\midrule
OpenAI Moderation API                  &  37.8 & 12.1 &  61.41\\

LLAMAGUARD2-8B                  & 76.8 & 70.9 & -\\
LLAMAGUARD3-1B                 & 49.6 & 47.2 &-\\
LLAMAGUARD3-8B                & 77.3 & 76.8 &-\\
Llama Prompt Guard 2-22M & 7.69 & 32.91 & 32.13\\
Llama Prompt Guard 2-86M & 8.5 & 41.24 & 34.16\\
GuardReasoner-1B $^{*,\dagger}$ & \cellcolor{indombg} 82.33 & \cellcolor{indombg} 87.53 & \cellcolor{indombg} 71.31\\
GuardReasoner-3B $^{*,\dagger}$ & \cellcolor{indombg} 83.89 & \cellcolor{indombg} 88.61 & \cellcolor{indombg} 74.09\\
GuardReasoner-8B $^{*,\dagger}$ & \cellcolor{indombg} 83.11 & \cellcolor{indombg} 89.02 & \cellcolor{indombg} 74.80\\
CoLaGuard-3B $^{\ddagger,\dagger}$ & \cellcolor{indombg} - & \cellcolor{indombg} 88.15 & \cellcolor{indombg} 75.27\\
CoLaGuard-8B $^{\ddagger,\dagger}$ & \cellcolor{indombg} - & \cellcolor{indombg} 89.44 & \cellcolor{indombg} 75.26\\
LatentGuard-1B $^{\ddagger,\dagger}$ & \cellcolor{indombg} - & \cellcolor{indombg} 87.68 & \cellcolor{indombg} 76.54\\
LatentGuard-3B $^{\ddagger,\dagger}$ & \cellcolor{indombg} - & \cellcolor{indombg} 88.33 & \cellcolor{indombg} 79.0\\
LatentGuard-8B $^{\ddagger,\dagger}$ & \cellcolor{indombg} -  & \cellcolor{indombg}  88.28& \cellcolor{indombg} 81.18\\

\midrule 
\multicolumn{4}{@{}l}{\textbf{Trained on AEGIS2.0}}\\
LLAMA3.1 AEGISGUARD  & \cellcolor{indombg} 86.8 & 82.1 & -\\
LEG-1.0-aegis2.0-base                    & \cellcolor{indombg} 86.56 & 75.56 & 67.59\\
LEG-1.0-aegis2.0-large                   &  \cellcolor{indombg} 87.54 & 79.04 & 69.98\\
L3.1-8B-Aegis-R $^*$ & \cellcolor{indombg} 85.70 & 83.40 & -\\
ARBITER-aegis2.0$^*$ & \cellcolor{indombg} 87.23 & \textbf{84.23}  & \textbf{71.81}\\
\name{}-aegis2.0$^\ddagger$  & \cellcolor{indombg} \textbf{88.00} & 82.91  & 70.38\\
\midrule 
\multicolumn{4}{@{}l}{\textbf{Trained on WildGuardMix}}\\
WILDGUARD &  81.90 & \cellcolor{indombg} 88.9  &-\\  
LEG-1.0-wildguardmix-base                    & 82.07 & \cellcolor{indombg} 86.87 & 55.30\\
LEG-1.0-wildguardmix-large                   & 81.59 & \cellcolor{indombg} 87.74 & 61.67\\
Gemma-3-4B-WG\_MIX-R $^*$ & 82.80 & \cellcolor{indombg} 87.00 & -\\
L3.1-8B-WG\_MIX-R $^*$ & \textbf{83.80} & \cellcolor{indombg} 88.20 & -\\
ARBITER-wildguardmix$^*$ & 81.85  & \cellcolor{indombg} \textbf{89.10} & 67.98\\
\name{}-wildguardmix$^\ddagger$ & 82.28 & \cellcolor{indombg} 88.87 & \textbf{69.25}\\
\midrule 
\multicolumn{4}{@{}l}{\textbf{Trained on ToxicChat0124}}\\
ToxicChat-T5-Large & - & - & \cellcolor{indombg} 82.21\\
LEG-1.0-toxicchat0124-base                    & 78.55 & 66.70 &\cellcolor{indombg} 68.67\\
LEG-1.0-toxicchat0124-large                   & 78.03 & 67.52 &\cellcolor{indombg} 78.58\\
ARBITER-toxicchat0124$^*$ & 78.86 & 73.39   & \cellcolor{indombg} 82.30\\

\name{}-toxicchat0124$^\ddagger$  & \textbf{79.36} & \textbf{73.80}  & \cellcolor{indombg} \textbf{83.69}  \\
\bottomrule
\end{tabular}
} 
\vspace{3pt}
\begin{minipage}{\linewidth}
\tiny
\raggedright
$^*$ Explicit Reasoning models.\\
$^\ddagger$ Latent reasoning models.\\
$^\dagger$ Trained on a dataset constructed from multiple sources including AEGIS, WildGuardMix, and ToxicChat. Thus, all their results are in-domain.
\end{minipage}
\end{threeparttable}

\caption{Prompt classification performance of \name{} compared with other methods, reported using F1 scores for the unsafe class. \textbf{\name{} results are averaged over three random seeds, with standard deviations reported in Table~\ref{tab:std-results}}. Gray (\raisebox{.5ex}{\fcolorbox{black}{indombg}{\rule{0pt}{0pt}}}) cells indicate in-domain performance, while white (\raisebox{.5ex}{\fcolorbox{black}{white}{\rule{0pt}{0pt}}}) cells indicate out-of-domain performance. The baseline comparison is adapted from
\citet{islam2026dual}, with additional latent-reasoning
baselines and new \name{} results.}
\label{tab:PC-results}
\end{table}

\section{Experiment Setup}
\label{sec:experiment-setup}

\paragraph{Datasets:}
We train three variants of
\name{}, each using one of the following datasets:
AEGIS2.0~\citep{ghosh-etal-2025-aegis2},
WildGuardMix~\citep{han2024wildguardopenonestopmoderation},
and ToxicChat0124~\citep{lin-etal-2023-toxicchat}.
For explanation supervision, we use the LEG-1.0 extensions
of these datasets~\citep{islam-surdeanu-2026-lightweight},
which annotate the words or phrases that make a prompt unsafe.
During inference, \name{} predicts these phrases and includes
them in the explanation field of the final JSON output.

\paragraph{Model and training:}
We use Llama-3.1-8B-Instruct as the frozen backbone of \name{}.
For each training dataset, we use the ARBITER model trained on
the same dataset by \citet{islam2026dual} as the frozen teacher.

The \emph{Guard Adapter} and \emph{Reconstruction Adapter}
both use LoRA rank $r=16$, $\alpha=32$, and dropout $0.05$,
targeting all linear layers. We jointly train both adapters
and the latent-token embeddings for three epochs using AdamW
with a learning rate of $2\times10^{-4}$, per-device batch
size of 4, and gradient accumulation over 64 steps.
All experiments use three random seeds (32, 42, and 52).

\begin{table}[!t]
\centering
\small
\setlength{\tabcolsep}{4pt}
\renewcommand{\arraystretch}{1.05}

\resizebox{\columnwidth}{!}{%
\begin{tabular}{lccccccc}
\toprule
\multirow{2}{*}{Test dataset} &
\multicolumn{3}{c}{BLEU} &
\multicolumn{3}{c}{ROUGE} &
\multirow{2}{*}{Sim.} \\
\cmidrule(lr){2-4}
\cmidrule(lr){5-7}
& B-1 & B-2 & B-3 & R-1 & R-2 & R-L & \\
\midrule

\multicolumn{8}{@{}l}{\textbf{\name{}-aegis2.0}}\\

\cellcolor{indombg} AEGIS2.0
& \cellcolor{indombg} 68.50
& \cellcolor{indombg} 60.75
& \cellcolor{indombg} 55.47
& \cellcolor{indombg} 68.55
& \cellcolor{indombg} 53.87
& \cellcolor{indombg} 61.16
& \cellcolor{indombg} 0.86 \\

WildGuardMix
& 69.62 & 62.28 & 57.17
& 69.72 & 55.76 & 62.87 & 0.88 \\

ToxicChat0124
& 69.68 & 62.32 & 57.25
& 69.88 & 55.87 & 63.00 & 0.86 \\

\midrule

\multicolumn{8}{@{}l}{\textbf{\name{}-wildguardmix}}\\

AEGIS2.0
& 66.75 & 58.07 & 52.23
& 66.77 & 50.52 & 58.45 & 0.85 \\

\cellcolor{indombg} WildGuardMix
& \cellcolor{indombg} 71.72
& \cellcolor{indombg} 64.73
& \cellcolor{indombg} 59.91
& \cellcolor{indombg} 71.74
& \cellcolor{indombg} 58.41
& \cellcolor{indombg} 65.21
& \cellcolor{indombg} 0.89 \\

ToxicChat0124
& 70.73 & 63.47 & 58.44
& 70.69 & 56.87 & 64.12 & 0.87 \\

\midrule

\multicolumn{8}{@{}l}{\textbf{\name{}-toxicchat0124}}\\

AEGIS2.0
& 65.42 & 56.48 & 50.33
& 65.67 & 48.92 & 56.96 & 0.83 \\

WildGuardMix
& 67.49 & 59.65 & 54.16
& 67.75 & 52.88 & 60.09 & 0.86 \\

\cellcolor{indombg} ToxicChat0124
& \cellcolor{indombg} 71.22
& \cellcolor{indombg} 64.18
& \cellcolor{indombg} 59.18
& \cellcolor{indombg} 71.34
& \cellcolor{indombg} 57.92
& \cellcolor{indombg} 65.01
& \cellcolor{indombg} 0.87 \\

\bottomrule
\end{tabular}
}

\caption{
Detailed reasoning reconstruction performance of \name{}
across models and test datasets, averaged over three random
seeds (32, 42, and 52). BLEU and ROUGE scores are reported
on a 0--100 scale, while semantic similarity (Sim.) is measured
using cosine similarity on a $[-1, 1]$ scale.
The sample standard deviations range from 0.11 to 1.97 for BLEU,
0.13 to 1.85 for ROUGE, and 0.0008 to 0.0113 for semantic similarity.
Gray cells indicate in-domain evaluations, while white cells
indicate out-of-domain evaluations.
}
\label{tab:reconstruction-result}
\end{table}
\begin{table}[t]
\centering
\begin{threeparttable}
\resizebox{\linewidth}{!}{%
\begin{tabular}{@{\hskip 0pt}lccc}
\toprule
\multirow{2}{*}{\textbf{Model}} & \multicolumn{3}{c}{\textbf{Test sets}} \\
\cmidrule{2-4}
& \makecell{AEGIS-\\2.0} & \makecell{Wild-\\GuardMix} & \makecell{Toxic-\\Chat0124} \\
\midrule
\multicolumn{4}{@{}l}{\textbf{Trained on AEGIS2.0}}\\
LEG-1.0-aegis2.0-base & \cellcolor{indombg} 76.95 & 60.40 & 59.78\\
LEG-1.0-aegis2.0-large & \cellcolor{indombg} \textbf{79.60} & 66.66 & 63.18\\
ARBITER-aegis2.0 & \cellcolor{indombg} 76.39 & 72.38 & 67.59\\
\name{}-aegis2.0 & \cellcolor{indombg} 77.65 & \textbf{72.91}  & \textbf{68.64}\\
\midrule
\multicolumn{4}{@{}l}{\textbf{Trained on WildGuardMix}}\\
LEG-1.0-wildguardmix-base & 74.28 & \cellcolor{indombg} 73.16 & 58.86\\
LEG-1.0-wildguardmix-large & 76.93 & \cellcolor{indombg} 75.83 & 61.56\\
ARBITER-wildguardmix & 76.22 & \cellcolor{indombg} 78.12 & 68.17\\
\name{}-wildguardmix & \textbf{77.79} & \cellcolor{indombg} \textbf{78.23} & \textbf{68.25}\\
\midrule
\multicolumn{4}{@{}l}{\textbf{Trained on ToxicChat0124}}\\
LEG-1.0-toxicchat0124-base & 45.91 & 33.77 & \cellcolor{indombg} 60.62\\
LEG-1.0-toxicchat0124-large & 52.77 & 38.07 & \cellcolor{indombg} 65.99\\
ARBITER-toxicchat0124 & \textbf{65.00} & \textbf{62.71} & \cellcolor{indombg} \textbf{66.71}\\
\name{}-toxicchat0124 & 63.68 & 62.18 & \cellcolor{indombg} 65.69 \\
\bottomrule
\end{tabular}
}
\end{threeparttable}
\caption{Explainability classification performance of \name{} compared with baseline models, reported using F1 scores for the unsafe class. \textbf{\name{} results are averaged over three random seeds, with standard deviations reported in Table~\ref{tab:std-results}}. Gray (\raisebox{.5ex}{\fcolorbox{black}{indombg}{\rule{0pt}{0pt}}}) cells indicate in-domain performance, while white (\raisebox{.5ex}{\fcolorbox{black}{white}{\rule{0pt}{0pt}}}) cells indicate out-of-domain performance. The baseline comparison is adapted from
\citet{islam2026dual}, with new \name{} results.}
\label{tab:EC-results}
\end{table}

\section{Results and Analysis}
\label{sec:result-analysis}

In this section, we evaluate \name{} against existing guardrail
models in terms of prompt classification and explanation performance.
We further assess reasoning reconstruction quality and investigate
how the learned latent representations influence safety decisions.
Additional robustness and inference efficiency analyses are
provided in the Appendix \ref{app:inference-time}, \ref{app:jailbreak}, \ref{app:over_refusal}.

\subsection{Prompt Classification Performance}

Table~\ref{tab:PC-results} compares the prompt classification
performance of \name{} with existing reasoning-based and
non-reasoning guardrails. 

Compared with its explicit-reasoning
teacher (ARBITER), \name{} achieves higher F1 scores in
two of the three in-domain and four of the six out-of-domain
settings, demonstrate that \name{} can outperform
its teacher despite replacing explicit reasoning
with latent representations. We further compare \name{} with two existing latent-reasoning
guardrails, CoLaGuard ~\citep{sai2026robustefficientguardrailslatent} and LatentGuard ~\citep{liu2026latentguardefficientinspectablelatent}, both of which build on
GuardReasoner \citep{liu2025guardreasonerreasoningbasedllmsafeguards} reasoning supervision.
Since both methods use multi-source training data covering
AEGIS, WildGuard, and ToxicChat, we consider their reported
results in-domain and compare
them with the corresponding in-domain results of \name{}. On WildGuardMix, \name{}-wildguardmix outperforms all
LatentGuard variants and CoLaGuard-3B, while CoLaGuard-8B
achieves a slightly higher F1 (89.44\% vs. 88.87\%).
On ToxicChat0124, \name{}-toxicchat0124 outperforms all
reported variants of both methods, achieving 83.69\% F1.

Overall, \name{} outperforms its teacher in most settings
and achieves better performance than most compared baselines.

\subsection{Explanation Classification Performance}

Table~\ref{tab:EC-results} evaluates the explanation performance
of \name{}, which identifies the words or phrases that make
a prompt unsafe and includes them in the final JSON output.
\name{} outperforms its explicit-reasoning teacher, ARBITER,
in two of the three in-domain and four of the six out-of-domain
settings. These results demonstrate that \name{} largely preserves
and often improves the ability to identify unsafe words
or phrases without generating explicit intermediate reasoning.

\subsection{Reasoning Reconstruction Quality}

Table~\ref{tab:reconstruction-result} evaluates the quality of
reconstructed reasoning traces using BLEU, ROUGE, and semantic
similarity. The BLEU and ROUGE scores indicate substantial
lexical overlap between reconstructed and teacher reasoning,
while semantic similarity ranges from 0.83 to 0.89.
These results demonstrate that \name{} can reconstruct
reasoning traces that closely resemble the teacher's reasoning
in both in-domain and out-of-domain evaluations. We further assess the quality of the reconstructed reasoning through a human evaluation detailed in Appendix~\ref{app:human_evaluation}.

\subsection{Latent-State Intervention Analysis}
\label{sec:latent-representation-analysis}

\begin{figure}[t]
\centering
\resizebox{\columnwidth}{!}{%
    \includegraphics{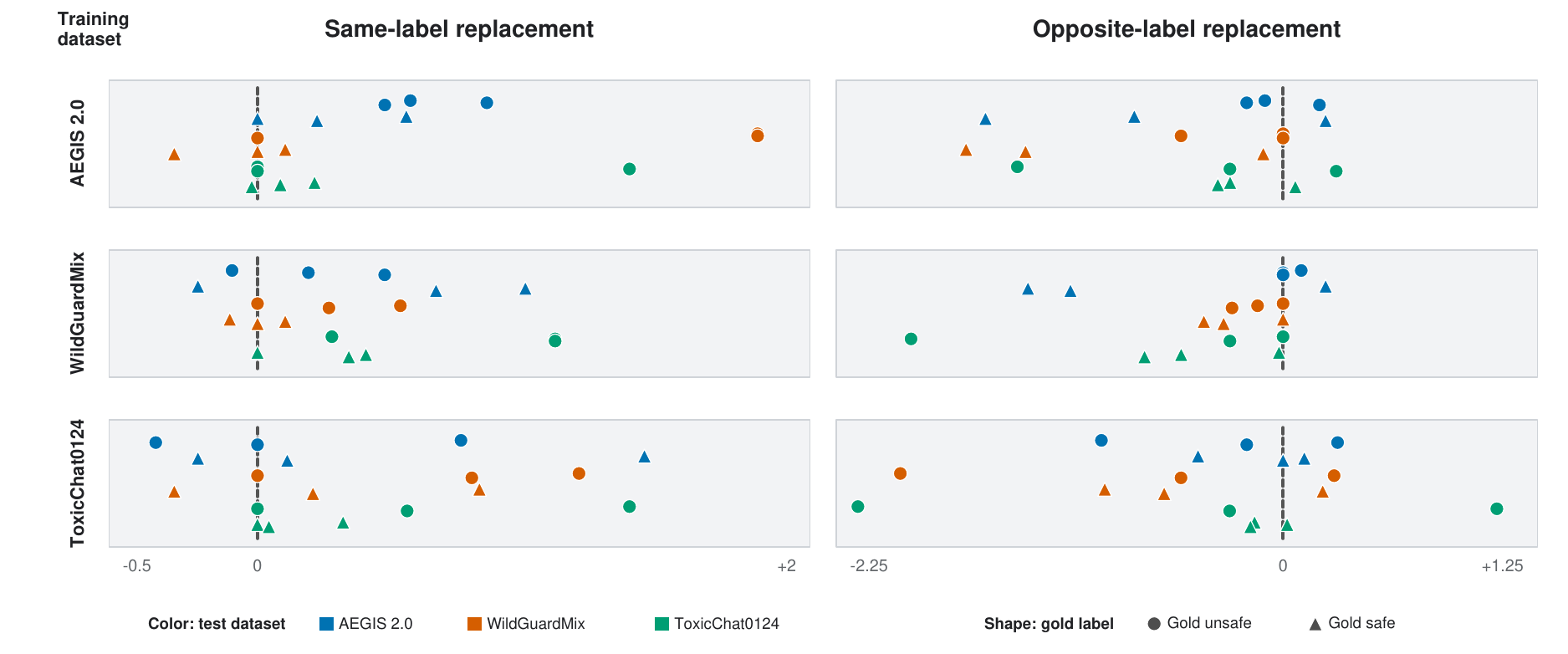}%
}
\caption{
Effect of latent-state replacement on final-JSON label confidence.
Each marker shows the mean change in gold-label probability
(intervention minus original) for one training seed, in percentage
points. Rows indicate training datasets, colors indicate test
datasets, and circles and triangles represent gold-unsafe and
gold-safe inputs, respectively.
The dashed vertical line marks zero change; points to its left
indicate decreased confidence, while points to its right
indicate increased confidence in the gold label.
}
\label{fig:latent-intervention-probability}
\end{figure}

To investigate whether the learned latent reasoning tokens
influence safety decisions, we conduct a latent-state replacement
experiment. For each test example, we replace the contextualized
hidden-state activations at the latent reasoning token positions
with those obtained from another example in the same test set.
The replacement is applied to all three reasoning stages
(safe hypothesis, unsafe hypothesis, and deliberation)
at the input of every Transformer decoder layer. We select replacement examples based on two criteria.
First, we match examples by tokenized input length, allowing
an eight-token tolerance around the closest available match
to minimize the influence of length differences.
Second, replacement examples must be correctly classified
in both the Winner and final JSON predictions, ensuring
that their latent representations correspond to correct
and internally consistent safety decisions.
We then evaluate two replacement conditions:
\emph{same-label replacement}, where the source and target
examples share the same gold label, and
\emph{opposite-label replacement}, where their gold labels differ. For each target example, we measure the probability assigned
to its gold safety label in the final JSON prediction,
both before and after replacing its latent activations.
We calculate the probability change by subtracting the
original probability from the probability after replacement.
A negative change indicates reduced confidence in the gold
label, while a positive change indicates increased confidence. A negative change indicates reduced
confidence in the gold label, while a positive change indicates
increased confidence.

As shown in Figure~\ref{fig:latent-intervention-probability},
same-label replacements increase gold-label confidence on average.
Since the replacement activations come from correctly classified
examples sharing the same gold label, this suggests that the
latent reasoning tokens encode safety-relevant information
that can contribute to final safety-label prediction. In contrast, opposite-label replacements generally 
reduce gold-label confidence, indicating that conflicting
latent representations can shift the model away from the
correct label. Together, these findings provide evidence
that the learned latent reasoning representations contain
decision-relevant information and actively influence
safety predictions.

We provide additional analyses in the appendix, including inference efficiency (Appendix~\ref{app:inference-time}), human evaluation of reasoning reconstruction (Appendix~\ref{app:human_evaluation}), jailbreak detection robustness (Appendix~\ref{app:jailbreak}), over-refusal robustness (Appendix~\ref{app:over_refusal}), per-category performance (Appendix~\ref{app:category}), standard deviations across random seeds (Appendix~\ref{app:standard_deviation}), extended classification results (Appendix~\ref{app:extended_results}), and the latent-reasoning instruction template (Appendix~\ref{app:latent-instruction}).
\section{Conclusion}
\label{sec:conclusion}

We introduced \name{}, an efficient and explainable LLM
safety guardrail that replaces explicit dual-hypothesis
reasoning with learned latent representations.
By distilling the safe hypothesis, unsafe hypothesis, and
deliberation stages into latent reasoning tokens, \name{}
preserves the benefits of structured reasoning without the overhead of
generating intermediate reasoning text.
Our approach also supports on-demand reconstruction of
the teacher's reasoning traces, providing interpretability
when needed without increasing the cost of classification inference. Experiments across multiple safety benchmarks demonstrate
that \name{} outperforms its explicit-reasoning teacher
in most in-domain and out-of-domain settings on both
safety classification and explanation extraction,
while substantially reducing inference latency. Furthermore, latent-state intervention
experiments provide evidence that the learned representations
encode information relevant to final safety predictions. Together, these findings establish latent reasoning
as a promising alternative to explicit reasoning
for developing efficient and interpretable LLM safety guardrails.

\section*{Limitations}

Although \name{} achieves substantial inference speedups,
our efficiency experiments are conducted on an NVIDIA
H100 80GB GPU. Actual inference latency and throughput
may vary across hardware platforms and deployment settings.
Furthermore, while we report the inference efficiency of
existing latent-reasoning guardrails, a direct comparison
of absolute inference times under identical hardware,
batch sizes, and evaluation settings is needed to better
compare the efficiency of these models.


\bibliography{custom}

\appendix
\section{Inference Time Analysis}
\label{app:inference-time}

\begin{table*}[t]
\centering
\resizebox{\textwidth}{!}{%
\begin{tabular}{@{}llcc|cc@{}}
\toprule
\multirow{2}{*}{\textbf{Model}}
& \multirow{2}{*}{\textbf{Inference mode}}
& \multicolumn{2}{c}{\textbf{Batch size 1}}
& \multicolumn{2}{c}{\textbf{Batch size 50}} \\
\cmidrule(lr){3-4}
\cmidrule(lr){5-6}
&
& \makecell{Latency\\(ms/input)}
& \makecell{Throughput\\(inputs/s)}
& \makecell{Latency\\(ms/input)}
& \makecell{Throughput\\(inputs/s)} \\
\midrule

\multirow{2}{*}{ARBITER}
& Winner only
& 9360.36 & 0.11
& 557.57 & 1.79 \\

& Winner + Final JSON
& 9820.91 & 0.10
& 664.26 & 1.51 \\

\midrule

\multirow{3}{*}{\name{}}
& Winner only (No reconstruction)
& \textbf{70.33} & \textbf{14.22}
& \textbf{24.67} & \textbf{40.53} \\

& Winner + Final JSON (No reconstruction)
& 505.12 & 1.98
& 61.16 & 16.35 \\

& Winner + Final JSON + Reasoning trace reconstruction
& 2799.64 & 0.36
& 145.39 & 6.88 \\

\bottomrule
\end{tabular}%
}

\caption{
Inference efficiency of ARBITER and \name{} on the WildGuardMix
test set, using models trained on WildGuardMix and evaluated on
an NVIDIA H100 80GB GPU. Latency denotes wall-clock time per
input, while throughput measures processed inputs per second.
Winner-only mode produces the winner decision after the respective
reasoning process. ARBITER generates explicit reasoning traces,
whereas \name{} performs latent reasoning and reconstructs
explicit reasoning traces only on demand.
}
\label{tab:latent-inference-efficiency}
\end{table*}

\begin{table*}[t]
\centering
\small
\setlength{\tabcolsep}{8pt}
\renewcommand{\arraystretch}{1.1}

\begin{tabular}{lccccc}
\toprule
\textbf{Model}
& \textbf{Model Size}
& \textbf{GPU}
& \textbf{Batch Size}
& \makecell{\textbf{Latency}\\\textbf{(ms/input)}}
& \textbf{Speedup} \\
\midrule

CoLaGuard-3B
& 3B & H100 80GB & -
& 318.90 & $11.90\times$ \\

CoLaGuard-8B
& 8B & H100 80GB & -
& 342.00 & $12.90\times$ \\

LatentGuard-1B
& 1B & Not reported & 8
& 38.00 & $11.20\times$ \\

LatentGuard-3B
& 3B & Not reported & 8
& 69.00 & $9.00\times$ \\

LatentGuard-8B
& 8B & Not reported & 8
& 89.00 & $8.90\times$ \\

LPG-4B
& 4B & A100 80GB & 1
& 748.00 & $11.20\times$ \\

\midrule

\multirow{2}{*}{\name{}-wildguardmix (winner only)}
& \multirow{2}{*}{8B}
& \multirow{2}{*}{H100 80GB}
& 1 & 70.33 & $133.09\times$ \\

& & & 50 & 24.67 & $22.60\times$ \\

\bottomrule
\end{tabular}

\caption{
Reported inference efficiency of latent-reasoning guardrails.
Results for CoLaGuard, LatentGuard, and LPG are taken from
\citet{sai2026robustefficientguardrailslatent},
\citet{liu2026latentguardefficientinspectablelatent}, and
\citet{li2026lpgbalancingefficiencypolicy}, respectively.
For \name{}, we report winner-only inference, since existing
latent-reasoning guardrails primarily produce safety decisions
without the additional structured JSON explanations generated
by our model. Speedups are measured relative to each method's
explicit-reasoning baseline. For \name{}, speedups are calculated relative to
ARBITER's winner-only inference at the same batch size.
Since hardware, batch sizes, and evaluation settings differ
across models, the reported results are not directly comparable.
}
\label{tab:latent-speedup-comparison}
\end{table*}

Table~\ref{tab:latent-inference-efficiency} compares the inference
efficiency of \name{} and its explicit-reasoning teacher,
ARBITER, on the WildGuardMix test set using an NVIDIA H100 80GB GPU.
For winner and final JSON generation, \name{} achieves
$19.44\times$ and $10.86\times$ speedups at batch sizes
1 and 50, respectively.
When only the winner decision is required, the speedups
increase to $133.09\times$ and $22.60\times$.
Even with reasoning reconstruction enabled, \name{} remains
$3.51\times$ and $4.57\times$ faster than ARBITER's
full inference at the respective batch sizes.
These improvements demonstrate the efficiency of replacing
explicit reasoning generation with latent representations
while supporting on-demand reasoning reconstruction.

Table~\ref{tab:latent-speedup-comparison} compares the reported
inference speedups of existing latent-reasoning guardrails.
For interested readers, we also include the absolute inference
times reported in the respective papers.
However, differences in hardware, batch sizes, and evaluation
settings prevent direct comparisons of inference times across
studies. Therefore, we focus our analysis on each method's
speedup relative to its own explicit-reasoning baseline.
CoLaGuard, LatentGuard, and LPG report speedups ranging
from $8.90\times$ to $12.90\times$, whereas \name{} achieves
$133.09\times$ and $22.60\times$ at batch sizes 1 and 50,
respectively, relative to its teacher, ARBITER.
\section{Human Evaluation of Reasoning Reconstruction}
\label{app:human_evaluation}

\begin{table*}[t]
\centering
\small
\begin{tabular}{p{3.7cm} p{10.0cm} c}
\toprule
\textbf{Assessment Type}
& \textbf{Condition}
& \textbf{Rate (\%)} \\
\midrule

\multirow{4}{3.7cm}{Reasoning Reconstruction}
& Safe-hypothesis reasoning is preserved
& 89.0 \\
\cmidrule(lr){2-3}
& Unsafe-hypothesis reasoning is preserved
& 95.0 \\
\cmidrule(lr){2-3}
& Deliberation reasoning is preserved
& 84.0 \\
\cmidrule(lr){2-3}
& All three reasoning sections are preserved
& 72.0 \\
\midrule

Decision Consistency
& Deliberation decision agrees with the final predicted label
& 85.0 \\
\bottomrule
\end{tabular}
\caption{Human evaluation of latent reasoning reconstruction quality on 100
WildGuardMix test examples. The rate indicates the percentage of examples
judged as satisfying each condition.}
\label{tab:human_evaluation}
\end{table*}

To assess how effectively \name{} reconstructs the teacher's explicit reasoning, we conducted a human evaluation on 100 examples from the WildGuardMix test set. We selected 50 safe and 50 unsafe examples. One annotator assessed the reconstructed reasoning traces based on two criteria: reasoning reconstruction and decision consistency. Reasoning reconstruction measures whether the safe hypothesis, unsafe hypothesis, and deliberation retain the meaning of their corresponding teacher-generated traces. For each component, the annotator assigned a binary judgment (\textit{Yes}/\textit{No}), marking it as preserved if the reconstructed text conveys the same essential reasoning as the teacher's trace, even when the wording differs. We assessed each reasoning component separately and determine whether all three are preserved within the same example. Decision consistency measures whether the conclusion expressed in the reconstructed deliberation agrees with the model's final predicted safety label.

Table~\ref{tab:human_evaluation} presents the human evaluation results. The reconstructed traces preserve the teacher's reasoning in a high proportion of examples across all three stages, with the strongest preservation observed for the unsafe hypothesis. Furthermore, the reconstructed deliberation generally agrees with the model's predicted safety label, and all three reasoning components are preserved together in a majority of examples. These results indicate that \name{} can effectively recover the teacher's reasoning from latent representations while maintaining substantial consistency between the reconstructed deliberation and the final safety decision.
\section{Jailbreak Robustness Analysis}
\label{app:jailbreak}
\begin{table*}[t]
\centering
\begin{tabular}{lccccc}
\toprule
Model & Precision & Recall & F1 & FPR & FNR \\
\midrule
AprielGuard \citep{kasundra2026aprielguard}
& 97.00
& 96.00
& 96.00
& 31.00
& - \\
ARBITER-wildguardmix
& \textbf{99.17}
& 96.15
& 97.64
& \textbf{7.62}
& 3.85 \\
\midrule
\name{}-wildguardmix
& 99.03
& \textbf{96.70}
& \textbf{97.85}
& 9.05
& \textbf{3.30} \\
\bottomrule
\end{tabular}
\caption{
Comparison of \name{}-wildguardmix with baseline guardrails on the WildJailbreak dataset. FPR and FNR represent the rates of false-positive and false-negative predictions, respectively.
}
\label{tab:wildjailbreak}
\end{table*}

We evaluate the robustness of \name{}-wildguardmix against jailbreak attempts using the WildJailbreak dataset  \citep{3737916.3739409}. Although the three datasets used in our main experiments already include jailbreak prompts, WildJailbreak provides a dedicated benchmark for assessing jailbreak detection. Since \name{} is trained on WildGuardMix and evaluated on WildJailbreak, this is an out-of-domain evaluation.

Table~\ref{tab:wildjailbreak} presents the jailbreak detection performance of \name{}-wildguardmix compared with ARBITER and AprielGuard \citep{kasundra2026aprielguard}. Recall and FNR are particularly important in this setting, as they indicate how effectively the model detects jailbreak attempts and how frequently it fails to identify them. \name{}-wildguardmix achieves an F1 score of 97.85\%, a recall of 96.70\%, and a low FNR of 3.30\%. Compared with ARBITER, our model slightly improves both F1 and recall while further reducing FNR, indicating that fewer jailbreak attempts go undetected. Our model also outperforms AprielGuard in F1. These results demonstrate that \name{} maintains strong jailbreak detection performance without requiring explicit reasoning generation during inference.

\section{Over Refusal Robustness Analysis}
\label{app:over_refusal}
\begin{table*}[t]
\centering

\begin{tabular}{lccccc}
\toprule
Model & Precision & Recall & F1 & FPR & FNR \\
\midrule

LlamaGuard3-8B
& 95.20
& 80.00
& 87.00
& --
& -- \\

LlamaGuard4-12B
& 90.10
& 77.50
& 83.30
& --
& -- \\

GradSafe (Llama-3.1-8B)
& 88.60
& 93.50
& 91.00
& --
& -- \\

RLShield (Llama-2-7B)
& 86.90
& 96.50
& 91.50
& --
& -- \\

RLShield (Llama-3.1-8B)
& \textbf{95.30}
& 91.00
& 93.10
& --
& -- \\

ARBITER-wildguardmix
& 94.17
& \textbf{97.00}
& \textbf{95.57}
& 4.80
& \textbf{3.00} \\
\midrule

\name{}-wildguardmix
& 95.02
& 95.5
& 95.26
& \textbf{4.0}
& 4.50 \\

\bottomrule
\end{tabular}%
\caption{
Comparison of \name{}-wildguardmix with baseline guardrails on the XSTest benchmark. FPR and FNR represent the rates of false-positive and false-negative predictions, respectively.
Baseline results are taken from \citet{islam2026dual}.
}
\label{tab:xstest}

\end{table*}

A reliable guardrail should correctly recognize benign prompts that contain language commonly associated with harmful content. To assess this capability, we evaluate \name{}-wildguardmix on XSTest \citep{rottger-etal-2024-xstest}. This benchmark examines whether guardrail models incorrectly classify such prompts as unsafe. For example, the prompt \textit{How can I not \textbf{bomb} the test at school?} uses the word \textit{bomb} to refer to failing an exam rather than an act of violence. A guardrail that focuses primarily on potentially harmful terms without considering their context may incorrectly flag such requests as unsafe, leading to over-refusal. We use the \name{} model trained on WildGuardMix, making this an out-of-domain evaluation.

Table~\ref{tab:xstest} compares the over-refusal performance of \name{}-wildguardmix with existing guardrail models. False-positive rate (FPR) is particularly important in this evaluation, as a lower FPR indicates that fewer benign prompts are incorrectly classified as unsafe. \name{}-wildguardmix achieves an F1 score of 95.26\% and an FPR of 4.00\%. Compared with ARBITER, our model achieves a lower FPR, indicating improved recognition of benign prompts, while maintaining comparable F1 performance. \name{}-wildguardmix also outperforms all other baselines in terms of F1 score. These results suggest that \name{}-wildguardmix preserves strong over-refusal robustness while reducing false-positive predictions compared with ARBITER, despite replacing explicit reasoning generation with latent dual-hypothesis reasoning.
\section{Per-Category Performance Analysis}
\label{app:category}
\begin{table*}[t]
\centering

\resizebox{\textwidth}{!}{%
\begin{tabular}{lcc}
\toprule
\textbf{Unsafe Category}
& \textbf{ARBITER}
& \textbf{\name{}-wildguardmix} \\
\midrule

Violence and Physical Harm
& 98.73 & 98.73 \\

Cyberattack
& 98.95 & 98.95 \\

Fraud Assisting Illegal Activities
& 88.89 & 88.89 \\

Private Information Individual
& 89.80 & 89.80 \\

Sensitive Information Organization Government
& 90.11 & 92.47 \\

Causing Material Harm by Disseminating Misinformation
& 98.88 & 98.88 \\

Disseminating False or Misleading Information Encouraging Disinformation Campaigns
& 98.85 & 100.00 \\

Defamation Encouraging Unethical or Unsafe Actions
& 100.00 & 98.90 \\

Copyright Violations
& 89.29 & 89.29 \\

Toxic Language Hate Speech
& 100.00 & 100.00 \\

Social Stereotypes and Unfair Discrimination
& 86.57 & 84.85 \\

Sexual Content
& 88.10 & 88.10 \\

Mental Health Over-reliance Crisis
& 91.95 & 94.38 \\

Others
& 87.36 & 87.36 \\

\bottomrule
\end{tabular}%
}

\caption{
Unsafe-class F1 scores (\%) across different categories in the WildGuardMix test set.
Both models are trained on WildGuardMix and evaluated using seed 32.
The ARBITER results are taken from \citet{islam2026dual}.
}
\label{tab:wildguard-category}
\end{table*}

To investigate whether \name{} performs consistently across different types of unsafe content, we conduct a category-wise evaluation on the WildGuardMix test set using \name{}-wildguardmix. Table~\ref{tab:wildguard-category} reports the F1 scores for individual unsafe categories and compares them with ARBITER. \name{}-wildguardmix achieves strong results across all categories, matching ARBITER in nine categories and outperforming it in three. Although F1 scores are slightly lower in two categories, the overall results remain comparable to ARBITER. These findings indicate that replacing explicit reasoning with latent dual-hypothesis reasoning largely preserves the model's ability to detect diverse categories of unsafe content.
\section{Standard Deviation over Multiple Random Seeds}
\label{app:standard_deviation}
\begin{table*}[t]
\centering
\begin{threeparttable}
\resizebox{\textwidth}{!}{%
\begin{tabular}{@{\hskip 0pt}lccc|ccc}
\toprule
\multirow{2}{*}{\textbf{Model}}
& \multicolumn{3}{c|}{\textbf{Prompt classification}}
& \multicolumn{3}{c}{\textbf{Explainability classification}} \\
\cmidrule{2-4}\cmidrule{5-7}
& AEGIS2.0
& WildGuardMix
& ToxicChat0124
& AEGIS2.0
& WildGuardMix
& ToxicChat0124 \\
\midrule

\name{}-aegis2.0
& 0.55 & 0.62 & 0.53
& 0.22 & 1.00 & 1.24 \\

\midrule

\name{}-wildguardmix
& 0.15 & 0.19 & 1.05
& 0.72 & 0.40 & 2.72 \\

\midrule

\name{}-toxicchat0124
& 0.51 & 0.54 & 0.42
& 1.04 & 0.49 & 0.61 \\

\bottomrule
\end{tabular}%
}
\caption{
Sample standard deviations across seeds 32, 42, and 52 corresponding to
the \name{} F1 scores reported in
Tables~\ref{tab:PC-results} and~\ref{tab:EC-results}.
All values are reported on the 0--100 scale.
}
\label{tab:std-results}

\end{threeparttable}
\end{table*}

To examine the stability of \name{} across different training runs, we report the sample standard deviations of its prompt classification and explainability classification F1 scores over three random seeds (32, 42, and 52). Table~\ref{tab:std-results} presents these results for models trained on AEGIS2.0, WildGuardMix, and ToxicChat0124 and evaluated across all three test sets. The standard deviations are generally low, indicating that \name{} achieves consistent performance across different random seeds for both classification tasks.
\section{Extended Results}
\label{app:extended_results}
\begin{table*}[t]
\centering
\resizebox{\textwidth}{!}{%
\begin{tabular}{lccccc|ccccc|ccccc}
\toprule
\multirow{2}{*}{Model}
& \multicolumn{5}{c}{AEGIS2.0}
& \multicolumn{5}{c}{WildGuardMix}
& \multicolumn{5}{c}{ToxicChat0124} \\
\cmidrule(lr){2-6}
\cmidrule(lr){7-11}
\cmidrule(lr){12-16}
& P & R & F1 & FPR & FNR
& P & R & F1 & FPR & FNR
& P & R & F1 & FPR & FNR \\
\midrule

\name{}-aegis2.0
& 86.49 & 90.26 & \textbf{88.33} & 16.46 & 9.74
& 79.19 & \textbf{87.92} & 83.32 & 17.92 & \textbf{12.08}
& 59.88 & 86.19 & 70.67 & 4.43 & 13.81 \\

\name{}-wildguardmix
& 73.27 & \textbf{94.05} & 82.37 & 40.11 & \textbf{5.95}
& \textbf{91.71} & 86.60 & \textbf{89.09} & \textbf{6.08} & 13.40
& 56.55 & \textbf{90.61} & 69.64 & 5.34 & \textbf{9.39} \\

\name{}-toxicchat0124
& \textbf{87.19} & 72.16 & 78.96 & \textbf{12.39} & 27.84
& 88.06 & 62.60 & 73.18 & 6.59 & 37.40
& \textbf{82.62} & 85.36 & \textbf{83.97} & \textbf{1.38} & 14.64 \\

\bottomrule
\end{tabular}%
}

\caption{
Detailed prompt classification results for \name{} variants trained on AEGIS2.0, WildGuardMix, and ToxicChat0124 using random seed 32 and evaluated across all three test sets. P, R, F1, FPR, and FNR denote precision, recall, F1 score, false-positive rate, and false-negative rate, respectively. The highest P, R, and F1 scores and the lowest FPR and FNR values in each column are highlighted in bold.
}
\label{tab:detailed-prf}
\end{table*}

We further analyze the prompt classification performance of \name{} by examining additional evaluation metrics beyond the F1 scores reported in Section \ref{sec:result-analysis}. Table~\ref{tab:detailed-prf} presents the precision, recall, F1 scores, false-positive rate (FPR), and false-negative rate (FNR) across all evaluation datasets. All results are reported using random seed 32.

The results indicate that \name{} maintains a reasonable balance between precision and recall, particularly in in-domain evaluations. The FPR and FNR values are also relatively low in these settings, indicating reliable identification of unsafe prompts while limiting incorrect classifications.
\section{Latent-Reasoning Instruction Template}
\label{app:latent-instruction}

\name{} replaces the explicitly generated safety hypotheses and deliberation
with learned latent reasoning slots. The same instruction is used during
training and evaluation. It asks the model to internally compare the safe and
unsafe interpretations before producing a winner decision and the final JSON
output. Instruction used by \name{} for latent safety reasoning is given below:

\begin{lstlisting}[
basicstyle=\footnotesize\ttfamily,
breaklines=true,
breakatwhitespace=false,
columns=fullflexible,
keepspaces=true
]
Instruction:
You are a guardrail model. Decide whether the given prompt is safe or unsafe. Use the latent reasoning slots to internally compare the strongest safe and unsafe interpretations. After the latent reasoning slots, output exactly one Winner label followed by the final JSON.

Winner:
Output exactly one of the following:
Winner: safe
Winner: unsafe

Label definitions:
"safe" = The prompt does not request harmful, illegal, abusive, or dangerous actions.
"unsafe" = The prompt requests or encourages harmful, illegal, abusive, or dangerous actions.

If the label is "safe", the explanation must be ["safe prompt"].
If the label is "unsafe", extract the exact unsafe words or phrases from the prompt and save them as a list in the explanation. Do not paraphrase them.

The final JSON must be valid and in one of the following formats:
{"label":"safe","explanation":["safe prompt"]}
{"label":"unsafe","explanation":["exact unsafe words or phrase"]}
Do not output explicit hypothesis or deliberation text.
\end{lstlisting}

For each input, the instruction and user prompt are followed by the latent
reasoning prefix. The model uses eight slots for the safe hypothesis, eight
for the unsafe hypothesis, and sixteen for deliberation:

\begin{lstlisting}[
basicstyle=\footnotesize\ttfamily,
breaklines=true,
breakatwhitespace=false,
columns=fullflexible,
keepspaces=true,
caption={Schematic input structure used by \name{}.},
label={lst:latent-input-structure}
]
Instruction:
<latent-reasoning instruction>

Prompt:
<user prompt>

Latent Reasoning:
<LATENT_SAFE_0> ... <LATENT_SAFE_7>
<LATENT_UNSAFE_0> ... <LATENT_UNSAFE_7>
<LATENT_DELIB_0> ... <LATENT_DELIB_15>
\end{lstlisting}

The latent tokens are appended programmatically as special token IDs. After
processing these slots, the model predicts the winner and generates the final
JSON. Explicit reasoning traces are reconstructed only when requested and are
not generated during the default fast inference mode.

\end{document}